\documentclass[10pt,twocolumn,letterpaper]{article}

\usepackage{cvpr}
\usepackage{times}
\usepackage{epsfig}
\usepackage{epstopdf}
\usepackage{float}
\usepackage{graphicx}
\usepackage{amsmath}
\usepackage{amsthm}
\usepackage{amssymb}
\usepackage{booktabs}
\usepackage{bm}
\usepackage[lined,boxed,ruled]{algorithm2e}
\usepackage{booktabs}
\usepackage{caption}
\def\bfC{{\boldsymbol{C}}}
\def\bfD{{\boldsymbol{D}}}

\def\bfH{{\boldsymbol{H}}}

\def\bfM{{\boldsymbol{M}}}

\def\bfP{{\boldsymbol{P}}}

\def\bfU{{\boldsymbol{U}}}
\def\bfV{{\boldsymbol{V}}}
\def\bfW{{\boldsymbol{W}}}
\def\bfX{{\boldsymbol{X}}}
\def\bfY{{\boldsymbol{Y}}}
\def\bfZ{{\boldsymbol{Z}}}

\usepackage[pagebackref=true,breaklinks=true,letterpaper=true,colorlinks,bookmarks=false]{hyperref}

\cvprfinalcopy 

\begin{document}

\title{ Multi-Image Semantic Matching by Mining Consistent Features}
\author{Qianqian Wang$^{\dag}$ \hspace{2em} Xiaowei Zhou$^{\dag}$ \hspace{2em} Kostas Daniilidis$^\ddag$ \vspace{1em}\\[0ex]
	$^\dag$ State Key Lab of CAD\&CG, Zhejiang University\hspace{0.5em}  \\$^\ddag$ GRASP Laboratory, University of Pennsylvania\\
}

\twocolumn[{%
\renewcommand\twocolumn[1][]{#1}%
\maketitle	
\begin{center}
    \centering
    \includegraphics[width=\textwidth]{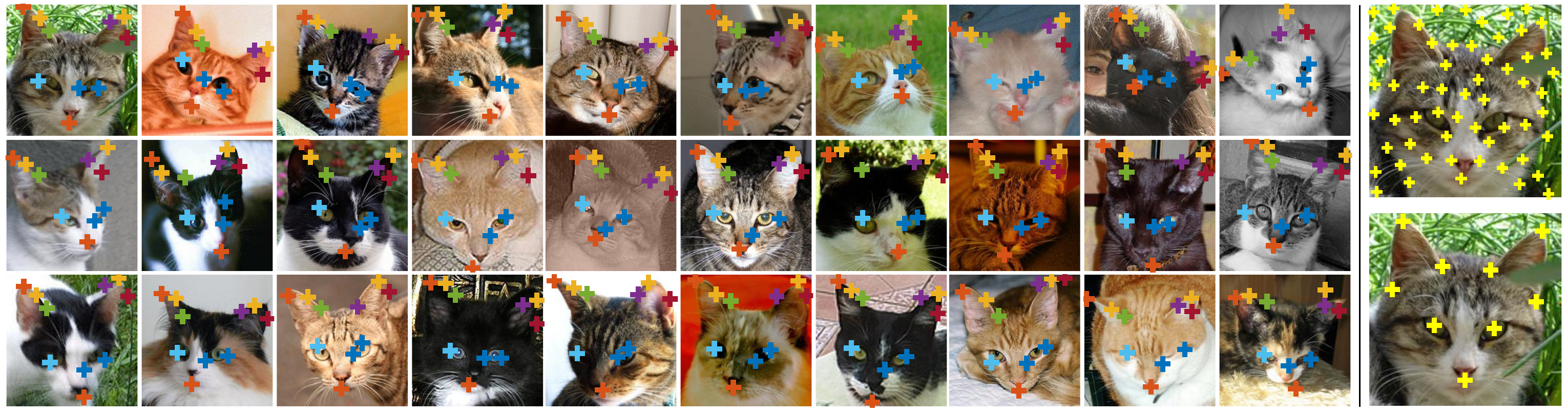}
    \captionof{figure}{Given initial feature candidates in multiple images and their noisy correspondences between each pair of images, the proposed method identifies a sparse set of reliable feature points and establishes cyclically and geometrically consistent correspondences across all images. The figure gives an example of identified features points (colored crosses) from 1000 cat head images (only 30 images are shown here). The color indicates the correspondence. The last column shows initial feature candidates (top) and manually-annotated landmarks provided by the dataset (bottom), both in the first image. Interestingly, the unsupervisedly-identified feature points by the proposed method roughly coincide with the manually-annotated landmarks.}\label{fig:catface}
\end{center}%
}]

\begin{abstract}
This work proposes a multi-image matching method to estimate semantic correspondences across multiple images.  In contrast to the previous methods that optimize all pairwise correspondences, the proposed method identifies and matches only a sparse set of reliable features in the image collection. In this way, the proposed method is able to prune nonrepeatable features and also highly scalable to handle thousands of images. We additionally propose a low-rank constraint to ensure the geometric consistency of feature correspondences over the whole image collection. Besides the competitive performance on multi-graph matching and semantic flow benchmarks, we also demonstrate the applicability of the proposed method for reconstructing object-class models and discovering object-class landmarks from images without using any annotation.

\end{abstract}
\section{Introduction}

Computing feature correspondences across images is a fundamental problem in computer vision. 
Low-level geometric features (e.g., SIFT \cite{lowe2004distinctive}) are successful for matching images of the same scene. Recently, there has been an increasing interest in semantic matching (e.g., \cite{liu2011sift,kim2013deformable}), i.e., establishing semantic correspondences across different object instances or scenes. Most research on semantic matching focuses on the pairwise case that considers only image pairs. Finding consistent correspondences across multiple images is important in many situations, e.g., object-class model reconstruction \cite{kar2015category} and automatic landmark annotation \cite{thewlis2017unsupervised}. The multi-image semantic matching problem is the focus of this work.

Despite remarkable advances in solving semantic matching and multi-image matching problems (see related work section), the following challenges remain. First, repeatable feature point detection for semantic matching is an open problem \cite{lenc2016learning,thewlis2017unsupervised}. Previous work bypassed this issue by either using all pixels (dense flow) \cite{liu2011sift} or randomly sampled points \cite{ufer2017deep}, resulting in numerous nonrepeatable features that have no real correspondences in other images. Second, previous multi-image matching methods (e.g., \cite{pachauri2013solving,zhou2015multi}) mainly optimize the cycle consistency of correspondences and seldom consider the geometric consistency simultaneously. While there have been effective ways to enforce geometric constraints in the pairwise setting (e.g., RANSAC \cite{fischler1981random} and graph matching \cite{leordeanu2005spectral}), few solutions exist for the multi-image case. Finally, the existing multi-image matching methods are computationally expensive, which could hardly process hundreds of images. Analyzing large datasets requires more scalable algorithms. 

In most situations one only needs the correspondences of a sparse set of highly repeatable features which are cyclically and geometrically consistent across images. Dense correspondences could be achieved by interpolation. Therefore, in contrast to the previous multi-image matching methods that optimize all pairwise correspondences, we formulate the problem as a feature selection and labeling problem: Starting from fussy pairwise correspondences, we aim to select a sparse set of feature points from the initial set of candidates in each image, and establish their correspondences across images by assigning labels to them. The selection and labeling are accomplished by optimizing both cycle consistency and geometric consistency of selected features. Formulating the problem in this way allows us to 1) explicitly deal with nonrepeatable feature points in the initial feature sets and 2) dramatically decrease the number of variables, resulting in a scalable algorithm that is able to jointly analyze thousands of images. Finally, inspired by classic results on factorization-based structure from motion \cite{tomasi1992shape}, we propose a low-rank constraint that enforces the geometric consistency for matching multiple images and is very efficient to solve in optimization. 
Figure \ref{fig:catface} gives an example illustrating our problem and the proposed method.

The main contributions of this work are summarized as follows:
\begin{itemize}
\item We propose a novel approach to solving the multi-image semantic matching problem as a feature selection and labeling problem. The proposed algorithm is able to discover consistent features in an image collection and is scalable to handle thousands of images.
\item We introduce a novel low-rank constraint for multi-image matching that allows the proposed algorithm to optimize cycle consistency and geometric consistency simultaneously.
\item We demonstrate the competitive performance of the proposed method on standard benchmarks. We also show two applications: 1) reconstruct 3D object-class models from images of different instances \emph{without using any manual annotation} and 2) match 1000 cat head images and interestingly find that the automatically selected feature points represent very discriminative landmarks on eyes, ears and mouths, which demonstrates the potential applicability of the proposed method to automatic landmark annotation.      
\end{itemize}

\section{Related work}

\noindent\textbf{Image matching}: In classical image matching, sparse feature correspondences between images are estimated using low-level geometric feature detectors (e.g., corners and covariant regions \cite{mikolajczyk2005comparison}) and descriptors (e.g., SIFT \cite{lowe2004distinctive}, SURF \cite{bay2006surf} and HoG \cite{horn1981determining}). The geometric consistency is imposed by either using RANSAC \cite{fischler1981random} as a postprocessing step or solving a graph matching problem that minimizes the geometric distortion between images \cite{leordeanu2005spectral,cho2010reweighted}. Many recent works attempt to find semantic correspondences across different scenes \cite{liu2011sift,revaud2016deepmatching}. Hierarchical matching \cite{kim2013deformable} and region-based strategies \cite{ham2017proposal} have been proposed to make use of high-level semantics in images. 

\noindent\textbf{Learning detectors and descriptors}: Recent results (e.g., \cite{fischer2014descriptor,long2014convnets}) show that the deep features extracted from convolutional neural networks (CNNs) are very effective in matching and outperform handcrafted features even if the CNNs are not trained for matching. Supervised learning has been used to explicitly learn descriptors. The supervision is from manually annotated correspondences \cite{choy2016universal}, images transformations \cite{novotny2017anchornet}, and additional cues and data, e.g., silhouettes \cite{kanazawa2016warpnet} and CAD models \cite{zhou2016learning}. Meanwhile, there are few efforts towards learning feature detectors that are repeatable and covariant to transformations \cite{yi2016lift,lenc2016learning,thewlis2017unsupervised}. The proposed method can be viewed as an unsupervised approach to harvesting reliable features and consistent correspondences from image collections, which may provide training data for detector and descriptor learning.

\noindent\textbf{Multi-image matching}: The proposed method is technically related to the joint matching methods \cite{kim2012exploring,huang2012optimization,pachauri2013solving}. Most existing methods aim to make use of the cycle consistency to improve the pairwise correspondences. Various approaches have been proposed such as unclosed cycle elimination \cite{zach2010disambiguating,nguyen2011optimization}, constrained local optimization \cite{yan2013joint,yan2014graduated,yan2015consistency,yan2015multi,zhou2015flowweb}, spectral relaxation \cite{kim2012exploring,huang2013consistent,pachauri2013solving} and convex relaxation \cite{huang2013consistent,chen2014near,zhou2015multi}. The proposed method differs from them as it aims to identify the most consistent features instead of optimizing all pairwise correspondences, making it more scalable for exploring large datasets. In addition, we introduce a low-rank constraint to optimize the geometric consistency of selected features. The matrix decomposition method for multi-graph matching proposed in \cite{yan2015matrix} imposes the low-rank constraint on graph edges. In this work the constraint is directly imposed on feature locations resulting in more efficient optimization.  The recent work \cite{tron2017fast} proposes an efficient method to discover clusters of discriminative features for matching, but no geometric constraint is considered.

\section{Preliminaries and notation}

\subsection{Pairwise matching}
Given $n$ images to match and $p_i$ feature points in each image $i$, the pairwise feature correspondences for each image pair $(i,j)$ can be represented by a partial permutation matrix $\bm P_{ij}\in\{0,1\}^{p_i\times p_j}$, which satisfies the doubly stochastic constraints:

\begin{equation}\label{eq:permutation1}
\bm{0}\le \bm P_{ij} \bm 1\le\bm{1}, \bm 0 \le\bm P_{ij}^T\bm{1} \le \bm{1}
\end{equation}

$\bm P_{ij}$ can be estimated by maximizing the inner product between itself and the feature similarities subject to the constraints in \eqref{eq:permutation1}. This is a linear assignment problem, which is well-studied and can be solved by the Hungarian algorithm. Finding $\bfP_{ij}$ can also be formulated as a graph matching problem, which can be cast as a quadratic assignment problem (QAP). Specifically, an objective function encoding both local compatibilities (feature similarity) and structural compatibilities (spatial rigidity) is maximized in order to find the assignment. Although QAP is NP-hard, many effective algorithms have been proposed to solve it approximately, e.g., \cite{berg2005shape,cho2010reweighted,leordeanu2005spectral}. We will use the output of linear matching or graph matching, denoted by $\bfW_{ij}\in \mathbb{R}^{p_i\times p_j}$, as our input.

\subsection{Cycle consistency} \label{sect:preliminaries cycle consistency}
Recent works \cite{chen2014near,pachauri2013solving,yan2014graduated} propose to use cycle consistency as a constraint to match multiple images. The correspondence between all pairs of images is cyclically consistent if the following equation holds for any triplet of images $(i,j,z)$:
\begin{equation}
\bfP_{ij} = \bfP_{iz}\bfP_{zj}
\end{equation}

The cycle consistency can be described more concisely by introducing a virtual ``universe'' which is defined as the set of unique features that appear in the image collection \cite{pachauri2013solving,huang2013consistent}. Each feature point in the universe must be observed by at least one image and matched to corresponding image points. Suppose the underlying correspondence between image $i$ and the universe is denoted by partial permutation matrix $\bfX_i\in \{0,1\}^{p_i\times u}$ , where $u$ is the size of the universe and $u \ge p_i$ for all $i$. The pairwise correspondence $\bfP_{ij}$ can be represented as $\bfX_i\bfX_j^T$. 

If the permutation matrices are concatenated as
\begin{equation}
\bm P = \begin{bmatrix}
\bm P_{11} & \bm P_{12} &\hdots & \bm P_{1n} \\
\bm P_{21} & \bm P_{22} &\hdots & \bm P_{2n} \\
\vdots & \vdots & \ddots & \vdots \\
\bm P_{n1} & \bm P_{n2} &\hdots & \bm P_{nn} \\
\end{bmatrix}, ~~
\bm X = \begin{bmatrix}
\bm X_{1} \\
\bm X_{2} \\
\vdots \\
\bm X_{n} \\
\end{bmatrix},
\end{equation}
it has been shown that the set $\{\bfP_{ij}|\forall i,j\}$ is cyclically consistent if and only if $\bfP$ can be factorized as $\bfX\bfX^T$ \cite{leordeanu2005spectral,huang2013consistent}.

\section{Proposed methods}

\subsection{Matching by labeling}\label{sect:Selection}
Recall that $\bfX \in \{0,1\}^{m\times u}$ is the map from image features to the universe, where $m$ and $u$ denote the total number of local features in the image collection and the size of universe, respectively. 
Another interpretation of $\bfX$ is that each row vector of $\bfX$ is the label of the corresponding feature. The image features with identical labels match each other. 
To accommodate all image features, previous work \cite{chen2014near,zhou2015multi} usually defines a sufficiently large $u$. 

However, not all features that appear in the image collection are desirable for matching. Particularly, in semantic matching most of the randomly or uniformly sampled features are nonrepeatable across images and should be excluded during matching. 
Inspired by this, we select the most repeatable image features and map them to a more compact feature space containing only $k$ elements, where $k$ is a predefined small value meaning the number of selected features in each image. 

Suppose the correspondences between the feature points in image $i$ and the selected feature space is represented by $\bfX_i\in\{0,1\}^{p_i\times k}$. Each $\bfX_i$ is a partial permutation matrix with a small number of columns which satisfies 
\begin{equation}
\bm{0}\le \bm X_{i} \bm 1\le\bm{1}, \bm X_{i}^T\bm{1} = \bm{1}
\label{eq:partialperm}
\end{equation}
The sum of each column in $\bfX_i$ equals to 1, meaning that every element in the selected feature space should correspond to a feature point in each image. On the contrary, the sum of a row could be zero, meaning that the corresponding feature point is not selected. 

The set $\{\bfX_i |1\leq i\leq n\}$ is what we need to estimate. $\bfX_i\bfX_j^T$ gives the pairwise correspondences between selected features in image $i$ and $j$, which must be cyclically consistent by construction as explained in Section \ref{sect:preliminaries cycle consistency}. As we attempt to identify the discriminative features that are supposed to produce more cyclically consistent correspondences in the initial pairwise matching, we minimize the discrepancy between the initial pairwise matching results and the constructed ones to estimate $\bfX$:
\begin{align}\label{eq:cycle consistency term}
\nonumber
\min_\bfX& \quad \frac{1}{4} \|\bfW - \bfX\bfX^T \|^2_F\\
\mbox{s.t.}& \quad \bfX_i \in \mathbb{P}^{p_i\times k}, 1 \le i \le n
\end{align}
where $\mathbb{P}$ denotes the set of partial permutation matrices and $\bfW \in \mathbb{R}^{m\times m}$ is the collection of $\bfW_{ij}$. By solving \eqref{eq:cycle consistency term}, the most repeatable features in the image collection will be selected and matched in a cyclically consistent way.

\subsection{Geometric constraint}\label{sect:Geometric}
Suppose we have tracked $k$ features over $n$ frames in a scene. 
We use $\bfM_i\in \mathbb{R}^{2\times k}$ to denote the coordinates of the $k$ ordered features of frame $i$, and concatenate all $\bfM_i$ as rows in a matrix $\bfM\in \mathbb{R}^{2n\times k}$ with each column per feature.
$\bfM$ is known as the \textit{measurement matrix} in structure from motion \cite{tomasi1992shape}. It can be shown that under orthographic projection, $\bfM$ is of rank $ 4 $.

In our problem, we represent the coordinates of all feature candidates in image $i$ by $\bfC_i\in\mathbb{R}^{2\times p_i}$. Then, the coordinates of selected feature points in image $i$ are given by
\begin{equation}
\tilde{\bfM_i} = \bfC_i\bfX_i
\end{equation}
where $\tilde{\bfM_i} \in \mathbb{R}^{2\times k}$ stores the coordinates of selected feature points that are aligned in the same order as the selected feature space. 
Similarly, we could concatenate all $\tilde{\bfM_i}$ as rows in a matrix $\tilde{\bfM}\in \mathbb{R}^{2n\times k}$:
\begin{equation}
\tilde{\bfM} =
\begin{bmatrix}
\bfC_1\bfX_1\\ \vdots \\ \bfC_n\bfX_n
\end{bmatrix}
\end{equation}

If feature points are correctly selected and labeled, $\tilde{\bfM}$ will become a measurement matrix of rank $ 4 $ under orthographic projection. Even if the scene is non-rigid, $\tilde{\bfM}$ can still be approximated by a low-rank matrix \cite{bregler2000recovering} . This conclusion can be effectively utilized to better estimate $\bfX$. Suppose the groundtruth rank of $\tilde{\bfM}$ is no larger than $r$. Minimizing the following term allows us to impose geometric consistency on selected feature points: 
\begin{align}
f_{\mbox{geo}} &=  \frac{1}{2}\|\tilde{\bfM} - \bfZ\|^2_F = \frac{1}{2}\sum_{i=1}^{n} \|\bfC_i\bfX_i-\bfZ_i\|^2_F
\label{eq: cost function of geometric consistency}
\end{align}
where $\bfZ\in \mathbb{R}^{2n\times k}$ is an auxiliary variable whose rank is no larger than $r$, and $\bfZ_i \in \mathbb{R}^{2\times k}$ denotes the $(2i-1)$-th and $2i$-th rows of $\bfZ$.

\subsection{Formulation}
Combining the cycle consistency and geometric consistency terms discussed in Section \ref{sect:Selection} and \ref{sect:Geometric}, we obtain the final optimization problem:
\begin{align}\label{eq:final objective function}
\nonumber
\min_{\bfX,\bfZ} &\quad \frac{1}{4}\|\bfW - \bfX\bfX^T \|^2_F + \frac{\lambda}{2}\sum_{i=1}^{n} \|\bfC_i\bfX_i-\bfZ_i\|^2_F\\
\mbox{s.t.} &\quad \bfX_i \in \mathbb{P}^{p_i\times k},1\le i\le n\\      	\nonumber
&\quad \mbox{rank}(\bfZ)\le r
\end{align}
where $\lambda$ controls the weight of the geometric constraint.

\subsection{Optimization}
We propose to solve the optimization problem in \eqref{eq:final objective function} by block coordinate descent, i.e., alternately updating one variable while fixing the others.

We attempted to relax the integer constraint of \eqref{eq:final objective function} and treated $\bfX$ as a real matrix $\bfX \in [0,1]^{m\times k}$, which is a common practice to solve quadratic assignment problems. However, we observed that, if the integer constraint on $\bfX$ is relaxed, the effect of the geometric constraint is negligible as the system $\bfC_i\bfX_i=\bfZ_i$ is ill-posed for arbitrary $\bfZ_i$. Therefore, we keep the integer constraint on $\bfX$. To make the optimization tractable, we decouple the two terms in \eqref{eq:final objective function} by replacing $\bfX$ in the first term with an auxiliary variable $\bfY \in \mathbb{R}^{m\times k}$ and rewrite the optimization as:
\begin{align}\label{eq:final optimization after relaxation}
	\begin{split}
		\min_{\bfX,\bfY,\bfZ} \ \ &\frac{1}{4}\|\bfW - \bfY\bfY^T\|^2_F + \frac{\lambda}{2}\sum_{i=1}^{n}\|\bfC_i\bfX_i - \bfZ_i\|^2_F\\
		&+ \frac{\rho}{2}\|\bfX - \bfY\|^2_F\\
		\mbox{s.t.} \ \  & \bfX_i \in \mathbb{P}^{p_i\times k}, 1\le i \le n\\ 
		\ \  & \bfY \in \mathcal{C}\\
		\ \  & \mbox{rank}(\bfZ) \le r
	\end{split}
\end{align}
where $\mathcal{C}$ denotes the set of matrices satisfying the following constraints:
\begin{align}
	\bm 0\le \bfY \le \bm 1, 
	\bm{0}\le \bfY_{i} \bm 1\le\bm{1}, \bfY_{i}^T\bm{1} = \bm{1}, 1\le i\le n
	\label{eq:relaxed constraints}
\end{align}
and $\rho$ is a parameter controlling the degree of similarity between $\bfX$ and $\bfY$. When $\rho$ approaches infinity, the problem in \eqref{eq:final optimization after relaxation} is equivalent to the original one \eqref{eq:final objective function}. 

The motivation for rewriting the problem as \eqref{eq:final optimization after relaxation} is that each subproblem in the block coordinate descent will be much easier to solve. We alternately update $\bfY$, $\bfX$, $\bfZ$ in the following manner.

$\bfY$ is updated via projected gradient descent \cite{parikh2013proximal,lu2016fast}:
\begin{align}\label{eq:update_Y_pg}
	\bfY \leftarrow \Pi_{\mathcal{C}}[\bfY - \eta(\bfY\bfY^T\bfY-\bfW\bfY+\rho(\bfY-\bfX))]
\end{align}
where $\Pi_{\mathcal{C}}$ denotes projection onto $\mathcal{C}$ and $\eta>0$ is the step-size. We update $\bfY$ until convergence before updating $\bfX$ and $\bfZ$.

Each $\bfX_i$ is updated via the Hungarian algorithm, whose cost matrix is constructed as
\begin{equation}\label{eq:Hungarian Cost Matrix}
	\bfH_{i} =  \lambda \bfD(\bfC_i,\bfZ_i) - 2\rho\bfY_i,
\end{equation}
where $\bfD(\bfC_i,\bfZ_i) \in \mathbb{R}^{p_i\times k}$ denotes the squared Euclidean distance between each pair of observations in $\bfC_i$ and $\bfZ_i$.

$\bfZ$ is updated via singular value decomposition (SVD):
\begin{equation}\label{eq:update_Z}
	\bfZ = \bfU \tilde{\bf\Sigma} \bfV^T,
\end{equation}
where the columns of $\bfU$ and the columns of $\bfV$ are the left and right singular vectors of $\tilde{\bfM}$ respectively, and $\tilde{\bf\Sigma}$ is a diagonal matrix with the diagonal elements corresponding to the $r$ largest singular values of $\tilde{\bfM}$.

For better convergence, we use an increasing sequence of $\rho$ to enforce the geometric constraint gradually. For each value of $\rho$, we update the variables alternately until the objective in \eqref{eq:final optimization after relaxation} does not decrease. 
As each update never increases the objective, the local convergence is guaranteed. 
In our experiment, we set the sequence of $\rho$ as ($1,10,100$), $\lambda$ as $1$ and $r$ as $4$ empirically.

As the optimization is nonconvex and involves both continuous and discrete variables, a reliable initialization is necessary. We first initialize $\bfY$ by ignoring the geometric constraint and solving
\begin{align}\label{eq:cycle consistency term after relaxation}
	\nonumber
	\min_\bfY& \quad \frac{1}{4}\|\bfW - \bfY\bfY^T \|^2_F\\
	\mbox{s.t.}& \quad \bfY\in \mathcal{C}
\end{align}
using the projected gradient descent as \eqref{eq:update_Y_pg} with $\rho=0$. $\bfX$ can be initialized by discretizing $\bfY$.

\section{Experiments}

\subsection{Multi-graph matching} \label{sect: experiment :multi-graph WILLOW/CMU}
We first validate the effectiveness of the proposed optimization algorithm in the setting of multi-graph matching, where the feature locations are annotated but their correspondences need to be estimated. The matching accuracy is evaluated by the recall, which is defined as the number of true correspondences found by the algorithm divided by the number of groundtruth correspondences.

We use the CMU datasets and the WILLOW Object Class dataset for evaluation. The CMU datasets contain the hotel sequence (111 frames) and the house sequence (101 frames). SIFT descriptors \cite{lowe2004distinctive} are extracted at 30 feature point annotations provided by \cite{caetano2009learning} in each frame. The WILLOW Object Class dataset \cite{cho2013learning} provides images of five object classes (Car, Duck, Motorbike, Face, Winebottle) and 10 annotated points corresponding to several discriminative parts of each class. Each class contains at least 40 images with different instances.  
As the object appearance in each class varies greatly, geometric descriptors like SIFT can hardly work. Instead, we adopt the deep features extracted from pretrained convolutional neural networks, which have proven to be effective in previous work \cite{long2014convnets}. Specifically, each image is fed through the AlexNet \cite{krizhevsky2012imagenet} (pretrained on ImageNet \cite{deng2009imagenet}) and the feature map responses of Conv4 and Conv5 corresponding to each landmark are extracted and concatenated as its descriptor.
For both datasets, the initial pairwise correspondences are obtained from the linear matching solver Hungarian algorithm and then fed into the proposed algorithm. Three alternative methods with publicly available code are used as baselines: 
the spectral method \cite{pachauri2013solving}, MatchLift \cite{chen2014near} and MatchALS \cite{zhou2015multi}. For all methods, the size of universe is set as the number of annotations in each image.  
\begin{table}
	\centering
	\renewcommand{\arraystretch}{1.3}
	\resizebox{\linewidth}{!}{
	\begin{tabular}{ccccccccc}
		\toprule
		Dataset		& Input	& \cite{pachauri2013solving}	& \cite{chen2014near} & \cite{zhou2015multi} & Ours$^-$ & Ours &Input$^+$	&Ours$^+$\\ 
		\hline
		Hotel 		& 0.57 	& 0.53	& 0.64	& 0.58	& 0.63	& 0.90	& 0.85 & 1 \\ 
		House		& 0.74	& 0.74	& 0.79	& 0.75	& 0.79	& 0.93	& 0.95 & 1	\\ 
		Car			& 0.48	& 0.55	& 0.66	& 0.65	& 0.72	& 0.75	& 0.83 & 1 \\		
		Duck		& 0.43	& 0.59	& 0.56	& 0.56	& 0.63	& 0.77	& 0.65 & 0.88\\
		Face		& 0.86	& 0.92	& 0.93	& 0.94	& 0.95	& 0.95	& 0.99 & 1	\\
		Motorbike	& 0.30	& 0.25 	& 0.28	& 0.27	& 0.40	& 0.61	& 0.85 & 1	\\
		Winebottle	& 0.52	& 0.64	& 0.71	& 0.72	& 0.73	& 0.82  & 0.92 & 1	\\
		\bottomrule
	\end{tabular}
	}
	\caption{The recall rates on the CMU datasets and the WILLOW Object Class dataset. The proposed method is compared with spectral method \cite{pachauri2013solving}, MatchLift \cite{chen2014near} and MatchALS \cite{zhou2015multi}. Ours$^-$ represents the recall rates without the geometric constraint. Input$^+$ and Ours$^+$ represent the initial and optimized recall rates respectively when graph matching is applied to obtain the initial pairwise correspondences.}
	\label{tbl:willow/cmu accuracy}
\end{table}
\begin{figure}
	\centering
	\includegraphics[width=0.48\linewidth]{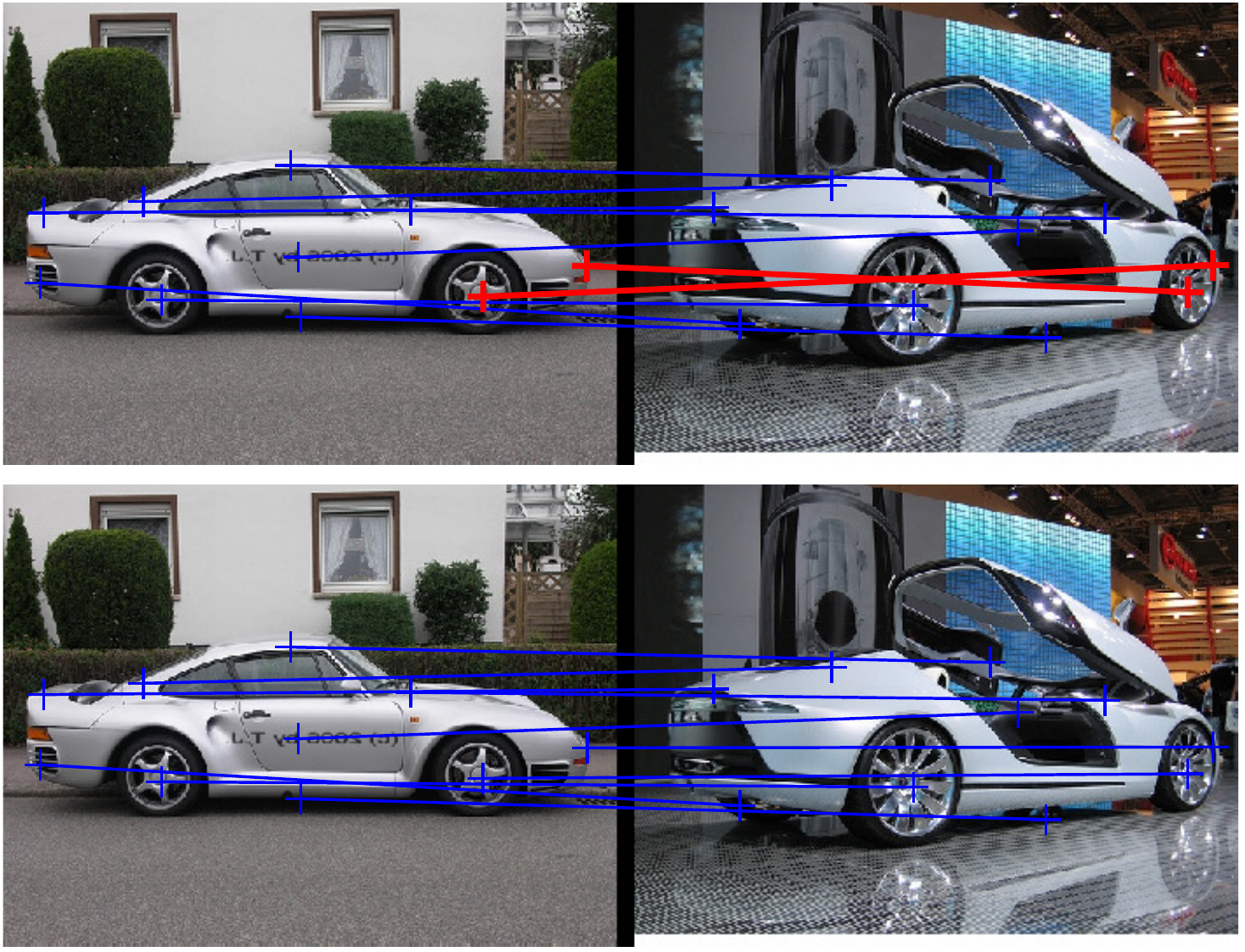}\hspace{0.02\linewidth}
	\includegraphics[width=0.48\linewidth]{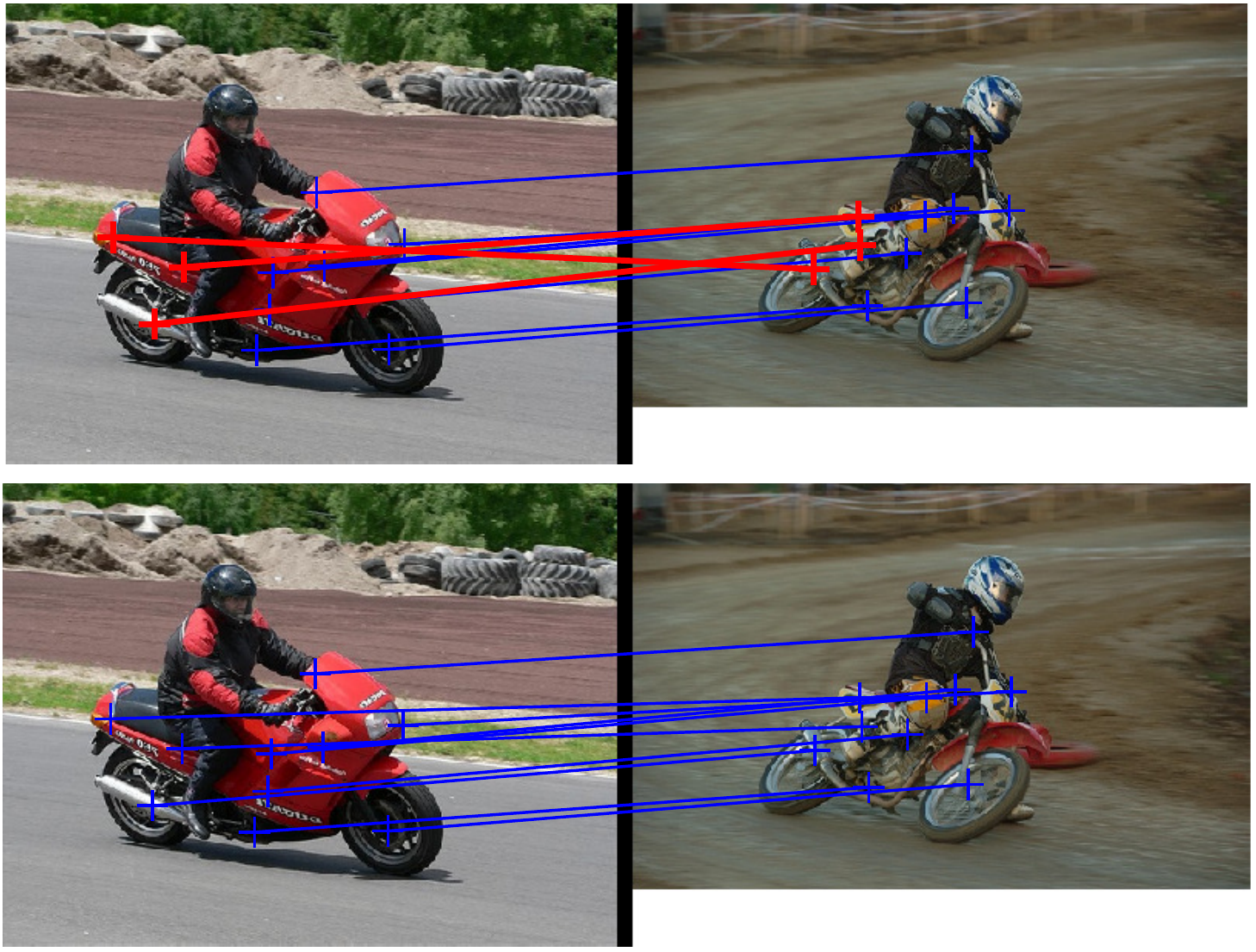}
	\caption{The matching results with and without the geometric constraint are shown in bottom and top rows, respectively. The true matches and false matches are shown in blue and red, respectively.}
	\label{fig:visualize geometric}
\end{figure}
\begin{figure}
	\centering
	\includegraphics[width=1\linewidth]{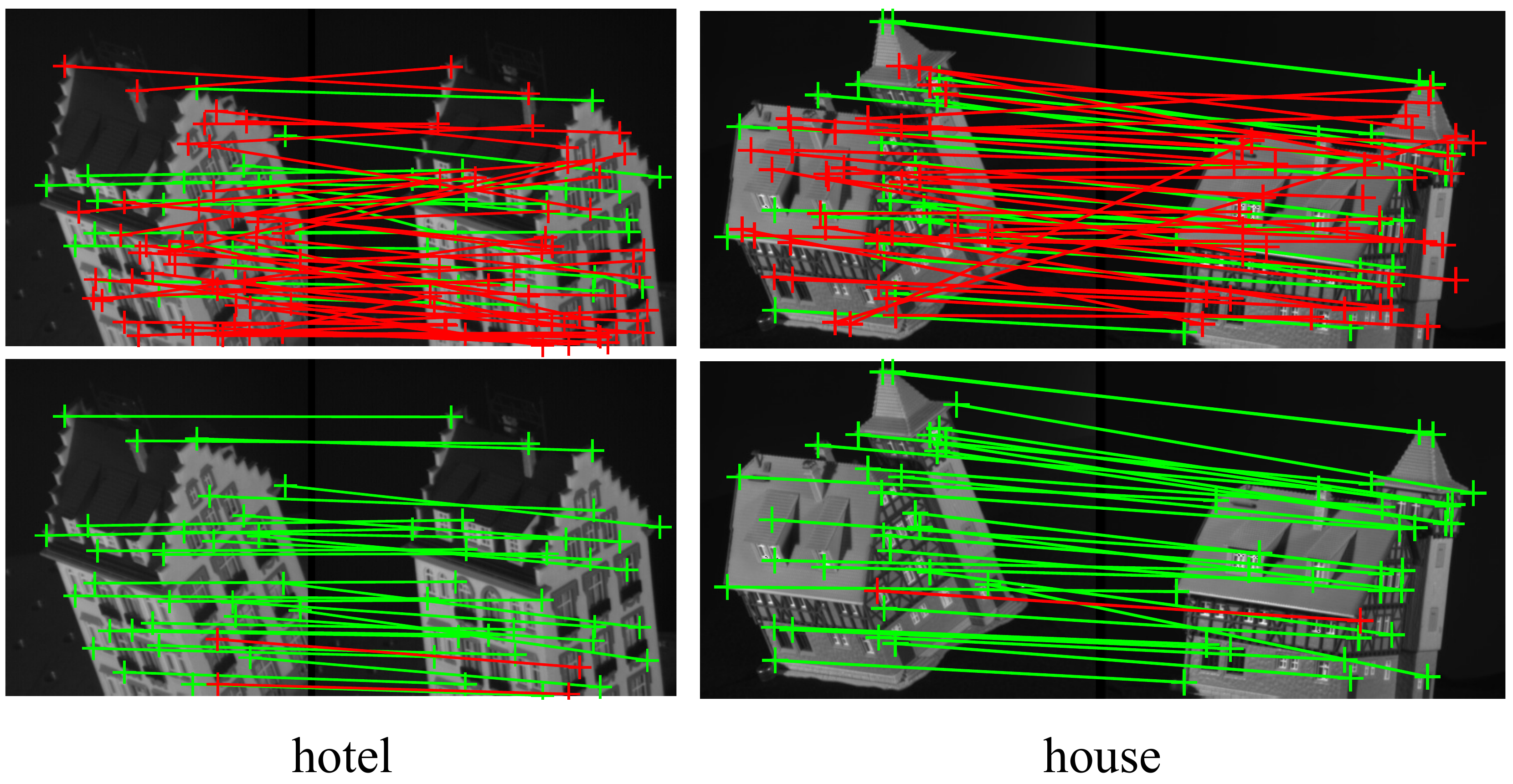}
	\caption{The ability to remove outliers. The true matches and false matches are shown in green and red, respectively. The top and bottom rows correspond to the results of pairwise matching and our joint matching method, respectively. Besides $30$ annotated feature points, we introduce $30$ randomly located points as outliers in each frame. We set $k = 30$ in our method.}
	\label{fig:visualize CMU outliers}	
\end{figure}
\begin{table*}\small
	\centering
	\renewcommand{\arraystretch}{1.2}
	\begin{tabular}{cccccccccccc}
		\toprule
		Methods		& car(S) & car(G) & car(M) & duc(S)	& mot(S) & mot(G) & mot(M) & win(w/o C) & win(w/C) & win(M) &Avg.\\
		\hline \hline
		LOM \cite{ham2017proposal} + Ours		
		& 0.89	 & 0.62	  & 0.56   & 0.70	& 0.49	 & 0.31	  & 0.28   & 0.91	    & 0.52	   & 0.72   & 0.60\\
		LOM	\cite{ham2017proposal}
		& 0.86	 & 0.58	  & 0.52   & 0.65	& 0.48	 & 0.28	  & 0.28   & 0.91		& 0.37	   & 0.65	& 0.56\\
		\hline
		DeepFlow \cite{revaud2016deepmatching}
		& 0.33	 & 0.13	  & 0.22   & 0.20	& 0.20	 & 0.08	  & 0.13   & 0.46		& 0.08	   & 0.18	& 0.20\\
		GMK \cite{duchenne2011graph}
		& 0.48	 & 0.25	  & 0.34   & 0.27   & 0.31   & 0.12   & 0.15   & 0.41       & 0.17     & 0.18	& 0.27\\
		SIFT Flow \cite{liu2011sift}
		& 0.54	 & 0.37	  & 0.36   & 0.32   & 0.41	 & 0.20   & 0.23   & 0.83  		& 0.16	   & 0.33	& 0.38\\
		DSP	\cite{kim2013deformable}
		& 0.46	 & 0.30	  & 0.32   & 0.25	& 0.31	 & 0.15	  & 0.14   & 0.85		& 0.25	   & 0.64	& 0.37\\
		Zhou \textit{et al.} \cite{zhou2016learning}
		& 0.77	 & 0.34	  & 0.52   & 0.42	& 0.34	 & 0.19	  & 0.20   & 0.78		& 0.19	   & 0.38   & 0.41\\
		\bottomrule
	\end{tabular}
	\caption{PCK ($\alpha = 0.1$) for dense flow on the PF-WILLOW dataset (SS w/HOG).
	}
	\label{tbl:comparison of semantic flow methods}	
\end{table*}

\begin{figure*}
	\centering
	\includegraphics[width=0.8\linewidth]{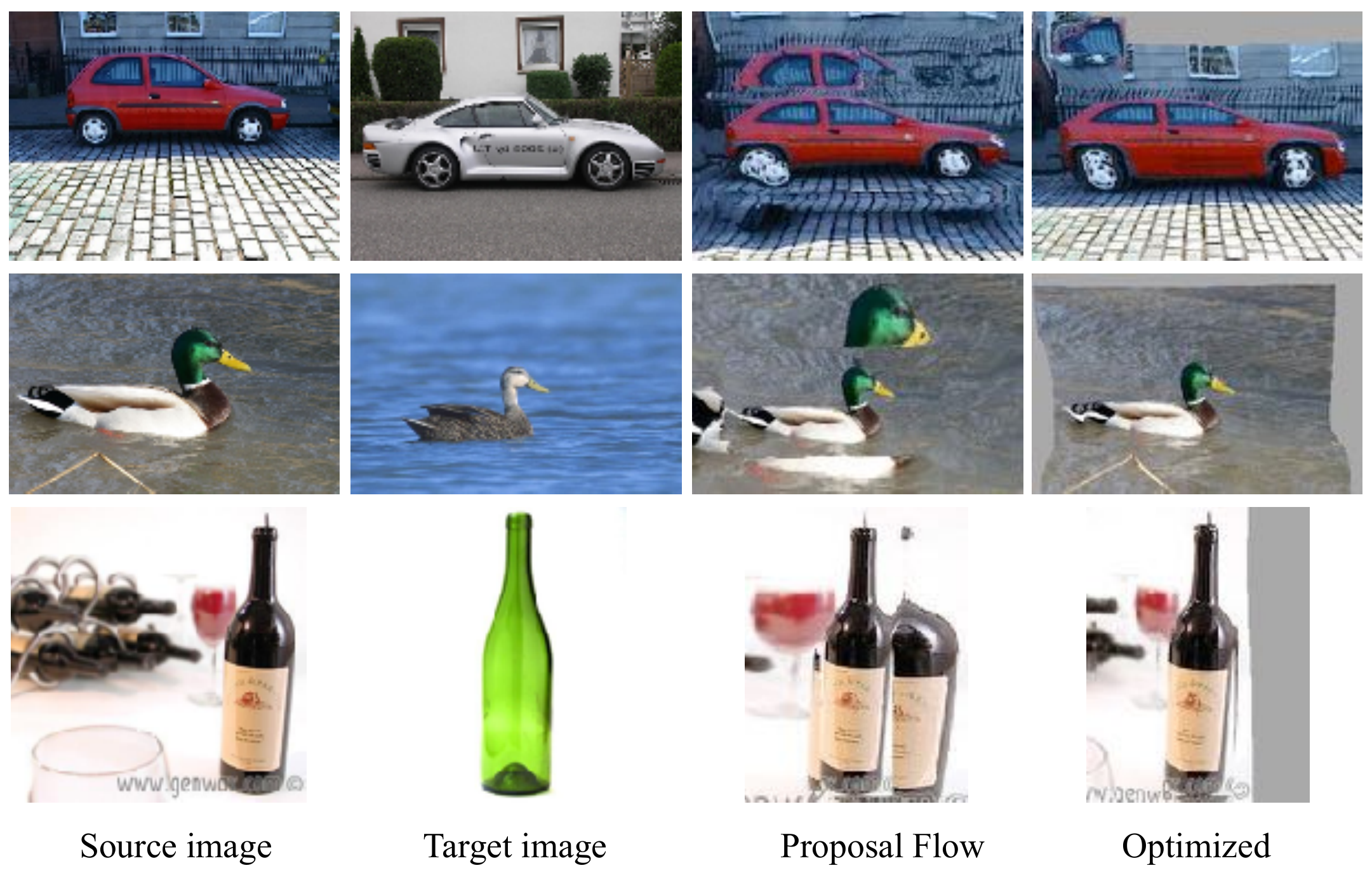}
	\caption{Examples of dense flow. The source images are warped to the target images using the dense correspondences estimated by proposal flow \cite{ham2017proposal} and optimized by the proposed method. }
	\label{fig: examples of dense flow}
\end{figure*}
The recall rates are reported in Table \ref{tbl:willow/cmu accuracy}, which shows that the proposed method outperforms other methods on all datasets. Table \ref{tbl:willow/cmu accuracy} also demonstrates another two cases: 1) If the graph matching solver RRWM \cite{cho2010reweighted} is leveraged to improve the initial pairwise correspondences, the matching accuracy of the proposed method can achieve $100\%$ on all datasets except the duck. 2) If the geometric consistency is ignored, the matching accuracy reduces significantly. Two sample image pairs are visualized in Figure \ref{fig:visualize geometric} which shows that geometrically distorted matches can be corrected after enforcing the geometric consistency. 

The proposed method can automatically select reliable features for matching. Figure \ref{fig:visualize CMU outliers} gives an example where randomly located feature points are added to images in the CMU datasets as outliers. It is shown that the outliers can effectively be pruned and moreover, the correspondences between the selected feature points are improved.
\subsection{Dense semantic matching}
In this section, we show the application of the proposed method in dense matching by combining it with region based semantic flow methods, e.g., proposal flow \cite{ham2017proposal}. In the proposal flow method, the correspondences of region proposals between images are estimated and then transformed into a dense flow field. For a collection of images, we apply the proposed method on top of proposal flow to improve the estimated pairwise correspondences of proposals, thus improving the dense flow. 

We experiment with a benchmark for evaluating semantic flow techniques named the PF-WILLOW dataset \cite{ham2017proposal}, which splits the WILLOW Object Class dataset into 10 sub-classes. They are car (S), (G), (M), duck (S), motorbike (S), (G), (M), winebottle (w/oC), (w/C), (M), where (S) and (G) represent side and general viewpoints, respectively, (C) denotes background clutter, and (M) denotes mixed viewpoints. Each sub-class includes 10 images of different object instances. 

The percentage of correct keypoints (PCK) is used as the evaluation metric \cite{ham2017proposal}. It evaluates the percentage of correctly located keypoints when transferring the annotated keypoints from an image to another image using the estimated flow. 
A predicted feature point is deemed to be correctly located if it lies within $\alpha$max$(h,w)$ pixels from the groundtruth point for $\alpha$ in $[0,1]$, where $h$ and $w$ are the height and width of the object bounding box, respectively. 
For proposal flow, the selective search (SS) \cite{uijlings2013selective} is used as proposal generator, HOG \cite{dalal2005histograms} as feature descriptors, and local offset matching (LOM) \cite{ham2017proposal} as geometric matching strategy. $500$ proposals are extracted in each image and used for matching and generating dense flow. In our algorithm, each proposal is treated as a feature point, and the center of each proposal is regarded as its coordinates in our geometric constraint. The number of selected features in all sub-classes is set as $10$.

The result in Table \ref{tbl:comparison of semantic flow methods} shows that our method improves the results of original proposal flow on most of the classes. A qualitative example is given in Figure \ref{fig: examples of dense flow}.

\subsection{Object-class model reconstruction}

Matching images of different object instances is a main challenge in reconstructing object-class models from images. Some previous works \cite{vicente2014reconstructing, carreira2015virtual,kar2015category} rely on annotated keypoints in images. The recent work \cite{zhou2015multi} requires no keypoint annotation but object masks to remove background. We show that the proposed method can produce consistent correspondences for reconstruction without using any manual annotation. We demonstrate with the FG3DCar dataset \cite{Lin2014jointly}, match all left-view sedan images ($37$ in total), and reconstruct a 3D model. 
In addition, we collect another dataset containing $30$ images of different motorbikes with similar views. 

Similar to \cite{zhou2015multi}, we uniformly sample feature candidates on image edges detected by the structured forests \cite{dollar2013structured}. Unlike \cite{zhou2015multi}, we don't need object masks thanks to the capability of the proposed method to prune nonrepeatable features in the background. On average, $\sim550$ feature candidates are obtained for each image. The deep features described in Section \ref{sect: experiment :multi-graph WILLOW/CMU} are used as descriptors and the graph matching solver RRWM \cite{cho2010reweighted} is adopted for initial pairwise matching. 

\begin{figure}
	\centering
	\includegraphics[width=0.7\linewidth]{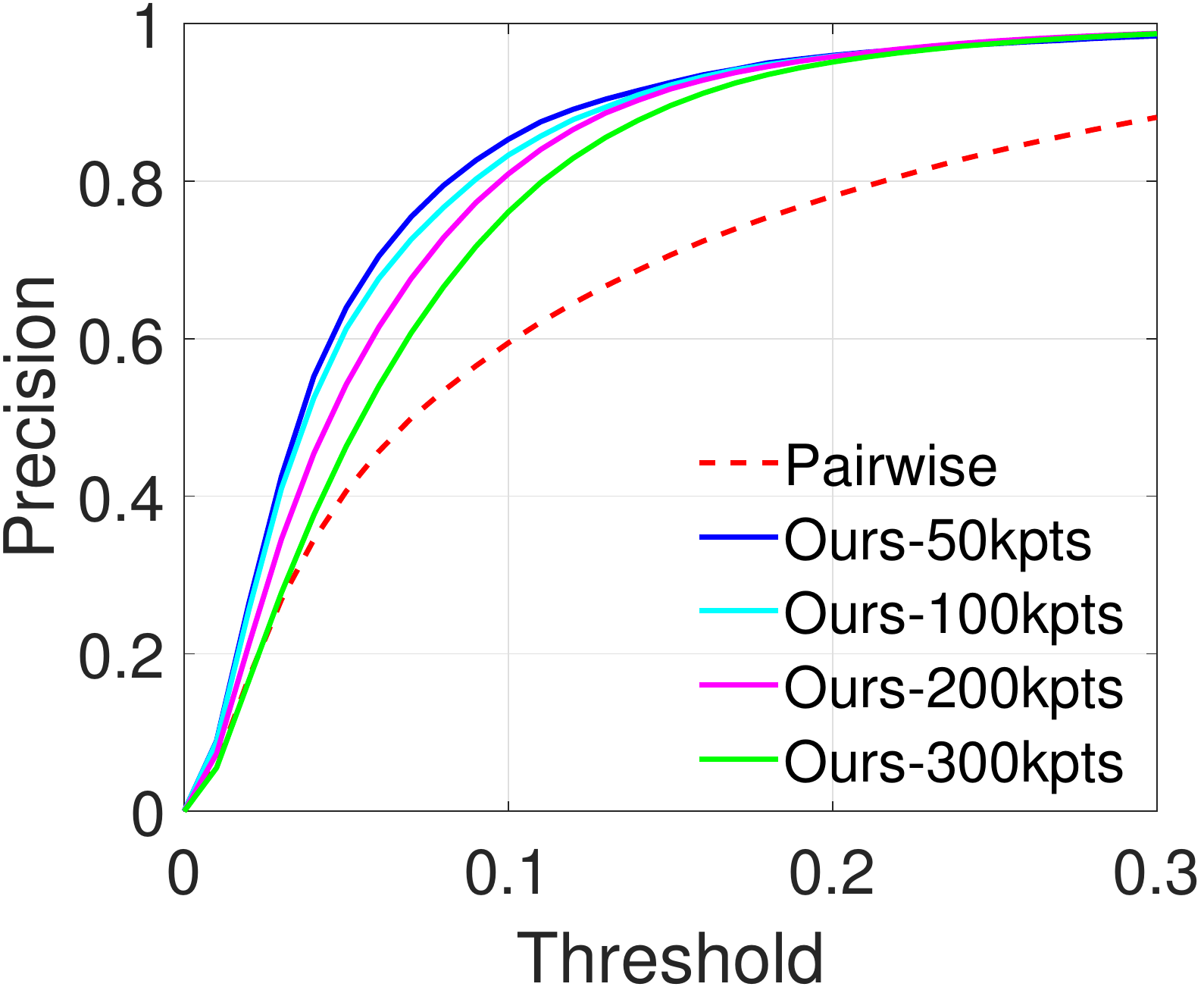}
	\caption{The precision for different numbers of selected features on the FG3DCar dataset.}\label{fig: Sedan_pck_50-300pts}
\end{figure}
\begin{figure}
	\centering
	\includegraphics[width=1\linewidth]{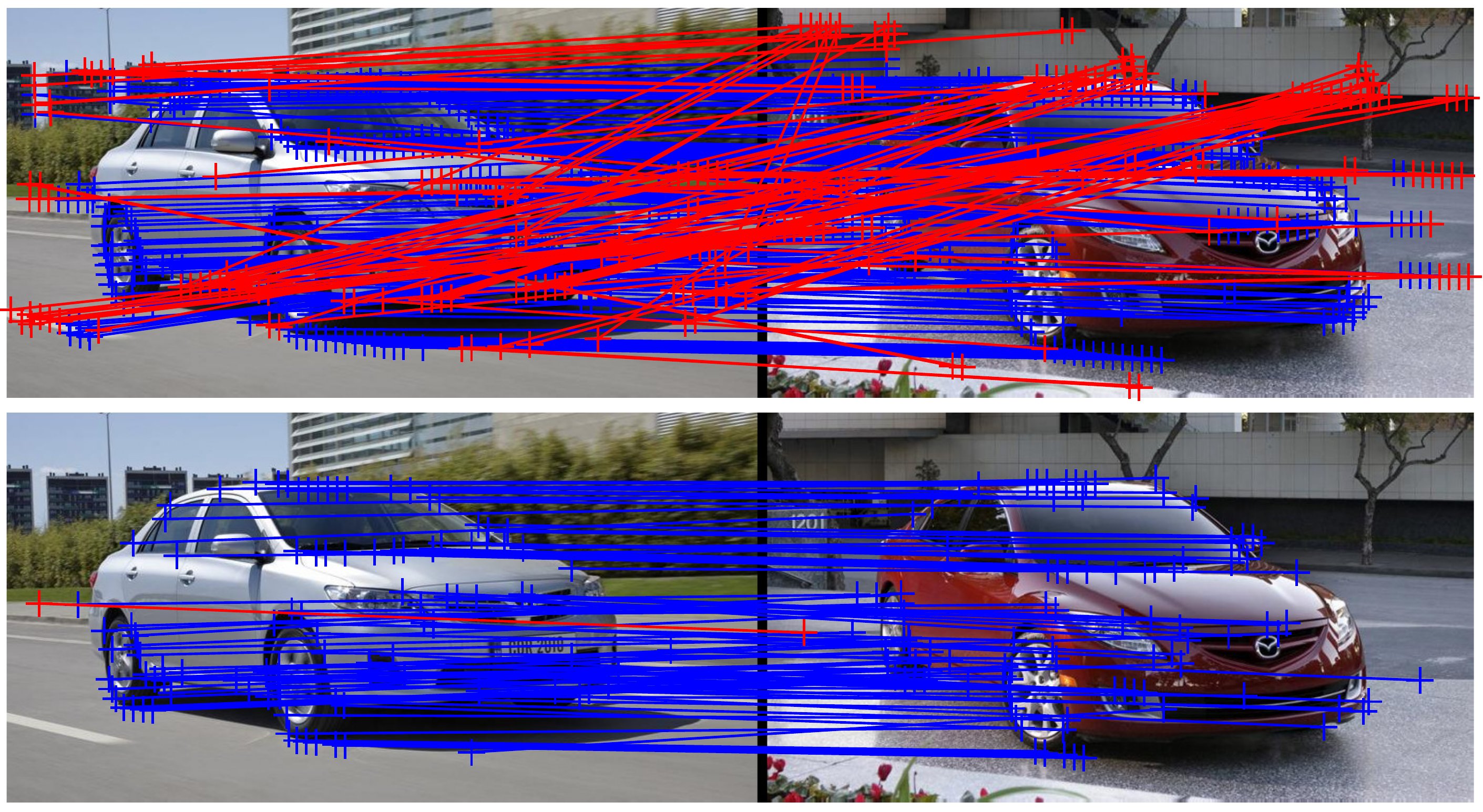}
	\caption{The matches between two sedan images. The true matches and false matches are shown in blue and red, respectively. The top and bottom rows correspond to the results of pairwise matching and the proposed method, respectively. Note that most of the initial feature points in the background are pruned by the proposed method. 
	}\label{fig: vis_FG3DCar_match_lines}
\end{figure}

To illustrate the effect of selection, we use the precision as our metric, which is defined as the number of true correspondences divided by the total number of correspondences found by the algorithm. The definition of true correspondence is similar to that in PCK. We vary the number of selected features and compare the precisions of recovered correspondences in Figure \ref{fig: Sedan_pck_50-300pts}.   
It is shown that the proposed method achieves obvious improvements compared to the original pairwise matching, whose precision is low in the presence of background clutter.  Moreover, the fewer the features we select, the higher the precision will be. This justifies the use of selection, which prunes nonrepeatable features such as the ones in the background. Only the reliable feature points near the objects are selected and matched consistently. An example is visualized in Figure \ref{fig: vis_FG3DCar_match_lines}. 

\begin{figure*}
	\centering
	\includegraphics[width=1\linewidth]{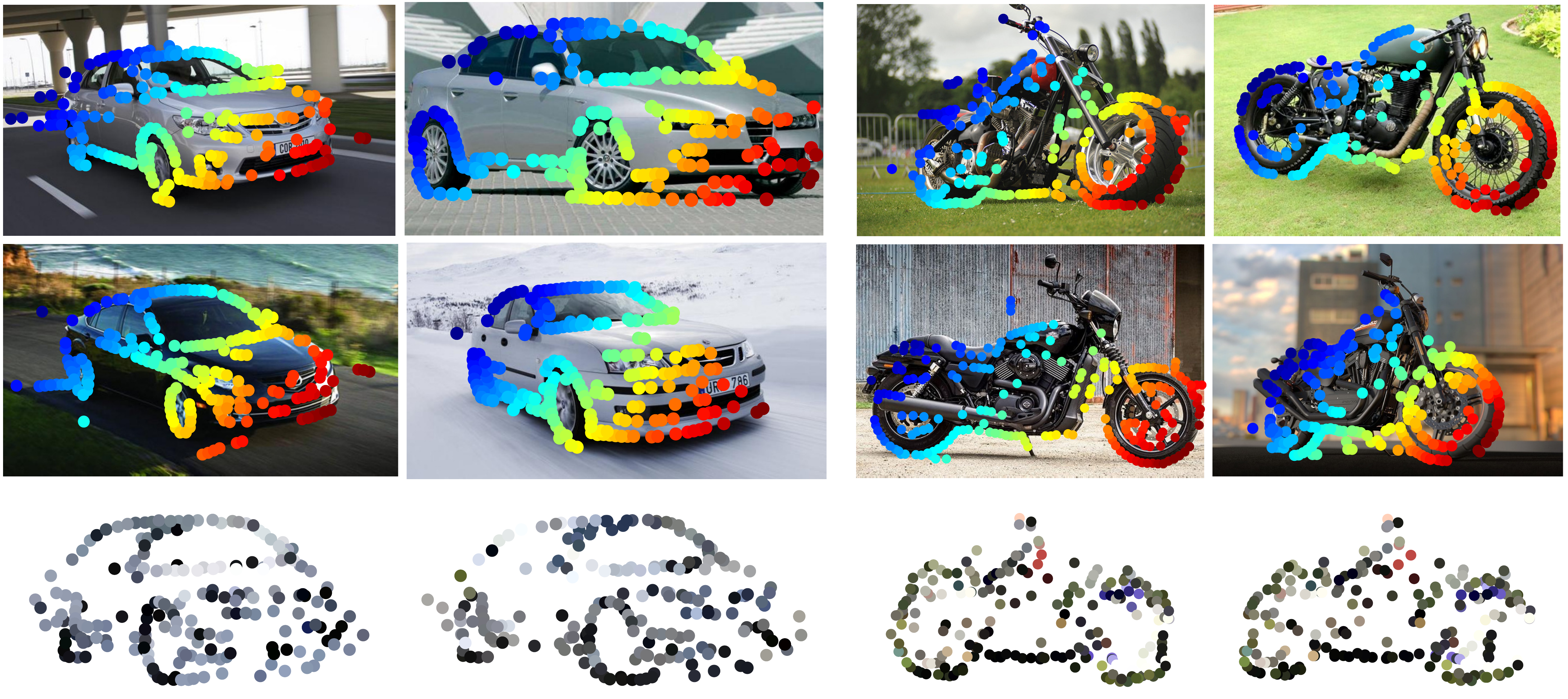}
	\caption{Matching sedans and motorbikes. Only four images are selected and shown for each image set. Note that the instances are different. All the feature points are automatically sampled and selected by the proposed method. The markers with the same color indicate the matched points. The 3D reconstruction is rendered with the colors in the first image and visualized in two viewpoints.}
	\label{fig:Matching sedans/motorbikes and reconstruction}
\end{figure*}

For reconstruction, we simply run affine reconstruction through the factorization method \cite{tomasi1992shape}. 
The selected feature points, correspondences and reconstructions are visualized in Figure \ref{fig:Matching sedans/motorbikes and reconstruction}. Clearly, most of the selected feature points are located on objects and correctly matched despite the large variety in object appearances and viewpoints. In spite of some noises and missing points, we can see the structures of the sedan and the motorbike from the reconstructions. It is believed that more sophisticated reconstruction techniques can be adopted to obtain better reconstructions. Quantitatively, we evaluate the estimated relative rotations between all pairs of images with the ground truth provided by the FG3DCar dataset. The mean error in geodesic distance is $18.5 ^\circ$. To our knowledge, no previous result exists for unsupervised relative pose estimation between different instances. 

\subsection{Automatic landmark annotation}\label{sect: experiment:unsupervised feature point annotation}
We apply the proposed algorithm to the first 1000 images from the cat head dataset \cite{zhang2008cat}. Similar to the previous experiment, the feature candidates are sampled from detected edges in images, yielding $\sim43$ candidates per image on average. We set the number of selected features to be $10$. The results are shown in Figure \ref{fig:catface}. As the figure shows, while initial candidates distribute randomly over the whole image including the background, the selected features are all on the objects with correct correspondences established across very different instances with a variety of appearances and poses. More interestingly, the automatically selected features roughly coincide with human annotations provided by the dataset, representing the discriminative parts of cat such as ears, eyes and mouth. This demonstrates the potential of the proposed method for automatic landmark annotation, which imitates humans' annotation process: we compare a collection of images and find a set of parts that are invariant in appearance and geometry across images. 

\subsection{Computational complexity}
The bottleneck restricting the scalability of the proposed method is the matrix multiplication when updating $\bfY$, while the other update steps only involve much smaller matrices. In \eqref{eq:update_Y_pg}, the dominant part is $\bfY\bfY^T\bfY$ which takes $O(mk^2)$ flops for each update, where $m$ and $k$ are the total number of features in all images and the number of selected features in each image, respectively. As a comparison, the complexity of MatchALS \cite{zhou2015multi} is $O(m^2k)$. As $k$ is much smaller than $m$, the proposed method is much more scalable. For example, $m\approx43,000$ and $k=10$ in the cat head experiment in Section \ref{sect: experiment:unsupervised feature point annotation}. We implement the proposed algorithm in Matlab on a PC with an Intel i7 3.4GHz CPU and 16G RAM. The CPU time for the cat head experiment is $\sim650s$ regardless of pairwise matching, which can hardly be solved by the previous multi-image matching algorithms. 

\section{Conclusion}

We presented a novel method that solved the problem of semantic matching across multiple images as a feature selection and labeling problem. The proposed method could establish reliable feature correspondences among a collection of images which satisfy both cycle consistency and geometric consistency. Experiments showed that the proposed method outperformed the previous multi-image matching methods while being highly scalable to match thousands of images. Several applications were demonstrated: improving dense flow estimation on top of the proposal flow method, reconstructing object-class models without using any manual annotation, and automatically annotating image landmarks in 1000 cat head images. 

\vspace{1em}
\noindent\textbf{Acknowledgements:} 
The authors are grateful for support through the following grants: ARL MAST-CTA
W911NF-08-2-0004, ARL RCTA W911NF-10-2-0016, ONR N00014-17-1-2093, and Fundamental Research Funds for the Central Universities (No. 2018FZA5011). 

\small
\bibliographystyle{ieee}
\bibliography{bibref}

\end{document}